\def\year{2022}\relax
\documentclass[letterpaper]{article} 
\usepackage{aaai22}  
\usepackage{times}  
\usepackage{helvet}  
\usepackage{courier}  
\usepackage[hyphens]{url}  
\usepackage{graphicx} 
\usepackage{natbib}  
\usepackage{caption} 
\DeclareCaptionStyle{ruled}{labelfont=normalfont,labelsep=colon,strut=off} 
\usepackage{algorithm}
\usepackage{algorithmic}

\usepackage{capt-of}
\usepackage{times}
\usepackage{epsfig}
\usepackage{graphicx}
\usepackage{amsmath}
\usepackage{amssymb}
\usepackage{xcolor}
\usepackage{blindtext}
\usepackage{overpic}
\usepackage{transparent}
\usepackage{booktabs}
\usepackage{multirow}
\usepackage[super]{nth}
\usepackage[format=plain,labelformat=simple,labelsep=period,font=small,skip=4pt,compatibility=false]{caption}
\usepackage[font=footnotesize,skip=2pt]{subcaption}
\usepackage{makecell}

\makeatletter
\newcommand\notsotiny{\@setfontsize\notsotiny\@vipt\@viipt}

\makeatother

\usepackage{etoolbox}
\usepackage[binary-units]{siunitx}
\robustify\bfseries
\DeclareSIUnit{\inch}{inch}

\newcolumntype{C}[1]{>{\centering\arraybackslash}p{#1}}

\usepackage{xspace}
\makeatletter
\DeclareRobustCommand\onedot{\futurelet\@let@token\@onedot}
\def\@onedot{\ifx\@let@token.\else.\null\fi\xspace}
\def\eg{\emph{e.g}\onedot} 
\def\ie{\emph{i.e}\onedot} 
\def\cf{\emph{c.f}\onedot} 
 \def\vs{\emph{vs}\onedot}
 
\def\etal{\emph{et al}\onedot}
\makeatother

\usepackage{pifont}
\definecolor{lightgray}{rgb}{0.9, 0.9, 0.9}
\definecolor{lgray}{rgb}{0.66, 0.66, 0.66}
\newcommand{\cmark}{\ding{51}\xspace}%
\newcommand{\xmarkg}{\textcolor{lgray}{\ding{55}}\xspace}%

\title{SCSNet: An Efficient Paradigm for Learning \\ Simultaneously Image Colorization and Super-Resolution}
\author{
   Jiangning Zhang$^1$\thanks{Work done during an intership at Tencent Youtu Lab.}
   ~ Chao Xu$^1$
   ~ Jian Li$^2$
   ~ Yue Han$^1$
   ~ Yabiao Wang$^2$
   ~ \textbf{Ying Tai}$^2$
   ~ \textbf{Yong Liu}$^1$\thanks{Corresponding author.} \\

   \textnormal{\normalsize $^1$APRIL Lab, Zhejiang University ~~ $^2$Youtu Lab, Tencent} \\
   {\tt\small \{186368, 21832066, 22132041\}@zju.edu.cn, yongliu@iipc.zju.edu.cn} \\
   {\tt\small \{swordli, caseywang, yingtai\}@tencent.com} \\
}

\begin{document}

\maketitle

\begin{abstract}
   In the practical application of restoring low-resolution gray-scale images, we generally need to run three separate processes of image colorization, super-resolution, and dows-sampling operation for the target device. However, this pipeline is redundant and inefficient for the independent processes, and some inner features could have been shared. 
   Therefore, we present an efficient paradigm to perform \textit{\textbf{S}}imultaneously Image \textit{\textbf{C}}olorization and \textit{\textbf{S}}uper-resolution (SCS) and propose an end-to-end SCSNet to achieve this goal. The proposed method consists of two parts: colorization branch for learning color information that employs the proposed plug-and-play \emph{Pyramid Valve Cross Attention} (PVCAttn) module to aggregate feature maps between source and reference images; and super-resolution branch for integrating color and texture information to predict target images, which uses the designed \emph{Continuous Pixel Mapping} (CPM) module to predict high-resolution images at continuous magnification. Furthermore, our SCSNet supports both automatic and referential modes that is more flexible for practical application. Abundant experiments demonstrate the superiority of our method for generating authentic images over state-of-the-art methods, \eg, averagely decreasing FID by 1.8$\downarrow$ and 5.1 $\downarrow$ compared with current best scores for automatic and referential modes, respectively, while owning fewer parameters (more than $\times$2$\downarrow$) and faster running speed (more than $\times$3$\uparrow$).
\end{abstract}

\begin{figure}[t]
   \centering
   \includegraphics[width=0.96\columnwidth]{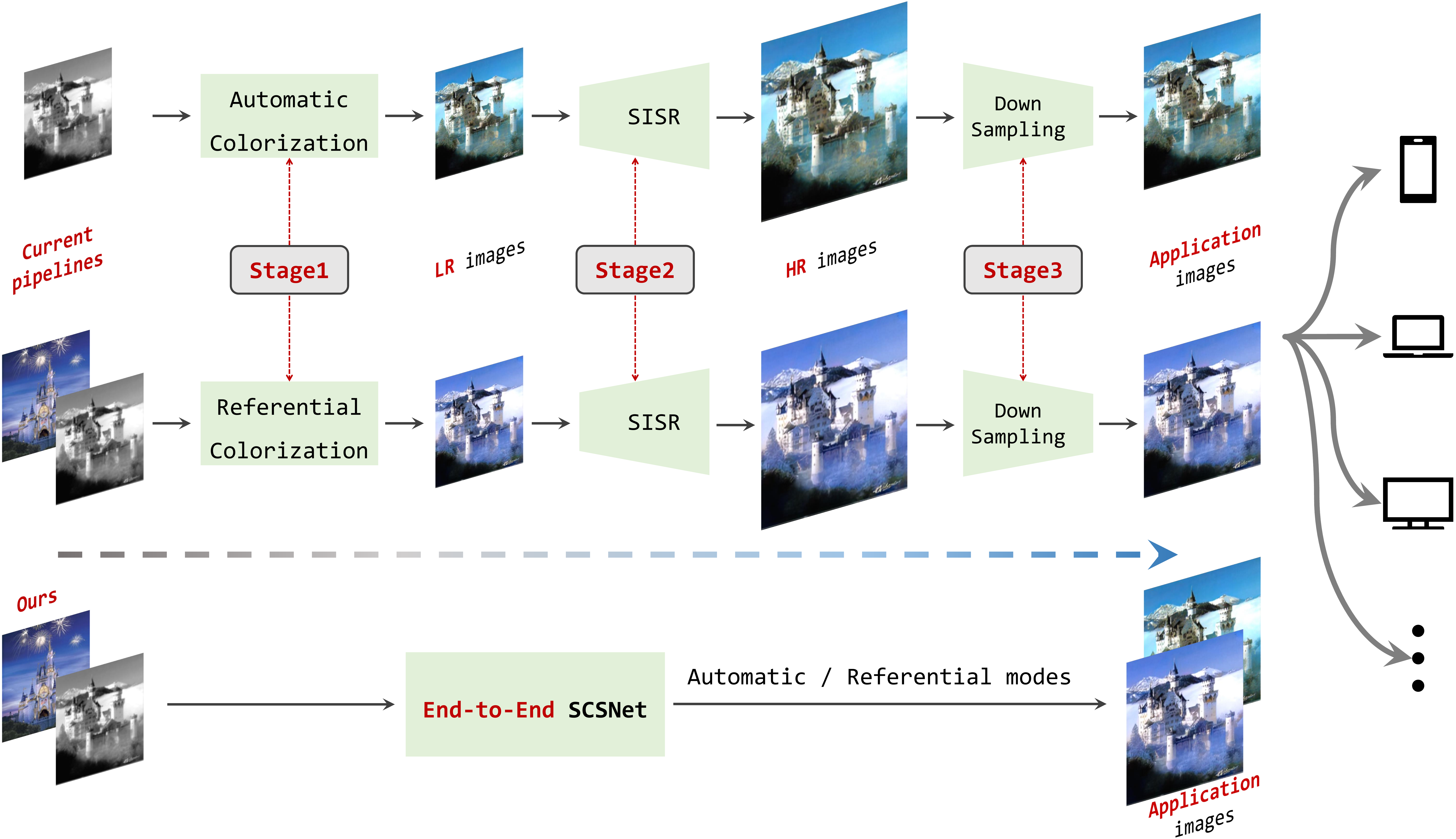}
   \caption{\textbf{Comparison between current pipelines and ours for the SCS task}. Current automatic and referential pipelines require three stages: 1) image colorization; 2) image super-resolution; and 3) down-sampling for target devices in different scenarios. Our end-to-end \emph{SCSNet} supports both automatic and referential modes.}
   \label{fig:Comparison}
\end{figure}

\begin{figure*}[t] 
   \centering
   \includegraphics[width=1.0\linewidth]{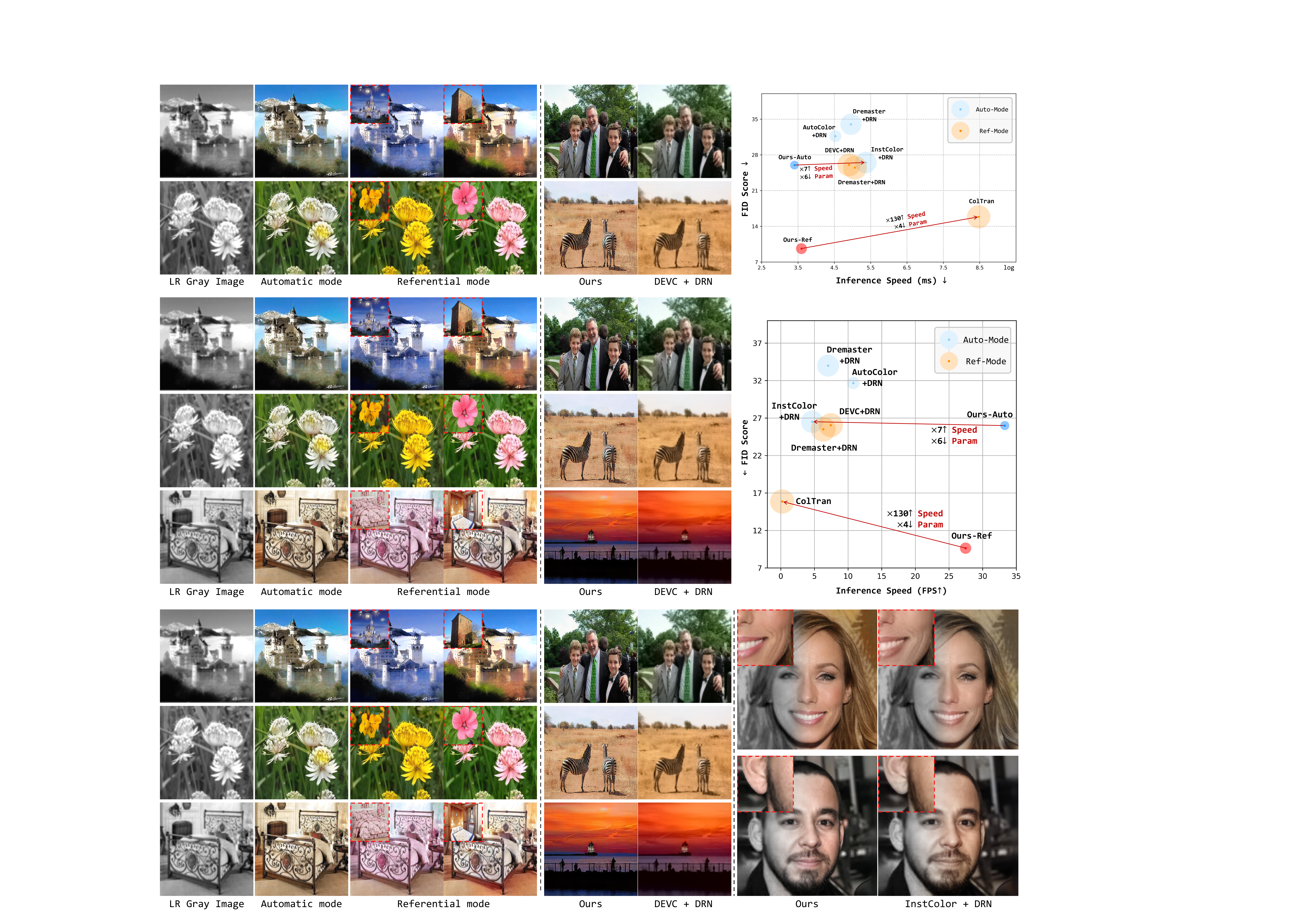}
   \caption{\textbf{Samples for simultaneously image colorization and super-resolution with 128$\times$128 gray inputs.} The left part shows our various results for 4$\times$ under \emph{automatic} and \emph{referential} modes, while the middle part is for 4$\times$ compared with SOTA referential pipeline. Right figure shows the efficiency comparison among our method and SOTA methods, and the circle size represents the parameter number of each method.}
   \label{fig:teaser}
\end{figure*}

\section{Introduction}
In some practical scenarios, \eg, restoration of old photos and artistic creation of gray-scale draft, we can only obtain Low-Resolution (LR) gray-scale images and hope to get more attractive High-Resolution (HR) colorful images. As shown in the top part of Figure~\ref{fig:Comparison}, the current solution pipeline cascades different methods by mainly three stages: 1) Using automatic or referential colorization model to color the gray-scale image for obtaining visually appealing RGB images. 2) Leveraging Single Image Super-Resolution (SISR) method for learning a nonlinear mapping to reconstruct HR images from LR inputs. 3) Down-sampling the generated HR images to the appropriate resolution for the target device. However, this pipeline is redundant and inefficient for practical use, where image colorization and super-resolution could have shared some common features by one unified network. Also, the device-adapted down-sampling operation in the last stage means that there is redundancy calculation in the SISR stage, which generates HR images at a fixed magnification (usually higher than needed) rather than device-required magnification. 
As shown in the bottom part of Figure~\ref{fig:Comparison}, we focus on solving the above problems and designing an efficient paradigm to achieve \emph{Simultaneously image Colorization and Super-resolution} (SCS) task by one unified network. Figure~\ref{fig:teaser} shows our authentic and diversified generation results in both automatic and referential modes, as well as the qualitative and quantitative comparison results with State-Of-The-Art (SOTA) pipelines. Concretely, we propose a novel efficient SCSNet that contains colorization and super-resolution branches.

\emph{For colorization branch, it learns how to predict two missing channels information from the given gray-scale image.} Image colorization mainly falls into automatic and referential modes depending on the availability of the reference image. The automatic mode only requires LR gray-scale image that seems intuitive but suffers from poor chromaticity of the generated images, because each semantic object can have various colors and the network tends to average output if applying inappropriate training strategy~\cite{auto_dl2}. The referential mode requires an additional reference image for providing semantic color information that is more controllable. A key point in the referential process is how to reasonably merge color information from the referential image into the source image. Some works~\cite{ref_5,ref_twostage2} propose to calculate the correlation matrix to characterize bidirectional similarity between source and reference images, and approaches~\cite{auto_seg2,ref_twostage1,auto_trans} use direct concatenation, AdaIN operation~\cite{adain}, or transformer module to aggregate information. However, the information interaction of current methods can be error-prone and may lack visual interpretation. Inspired by self-attention~\cite{sagan}, we redesign a plug-and-play \emph{Pyramid Valve Cross Attention} (PVCAttn) module that applies interpretable valves to control the information flow and fuses features at multiple scales. Also, our SCS paradigm supports both modes controlled by a reference switch in the PVCAttn module.

\emph{For super-resolution branch, it learns how to reconstruct HR images from LR images.} In general, SISR technology is employed to post-process the generated images for better visualization, and almost all current SISR methods only carry out fixed magnification~\cite{sisr_gan2,sisr_loop}, which goes against the natural world with a continuous visual expression. Unlike recent Meta-SR~\cite{metasr} that attempts continuous magnification by predicting convolutional weights for each pixel, we propose a more efficient \emph{Continuous Pixel Mapping} (CPM) module to realize arbitrary magnification in a continuous space. Specifically, we make the following three contributions:

\begin{itemize}
\item We propose an efficient SCSNet paradigm to perform the SCS task in an end-to-end manner firstly, and abundant experiments demonstrate the superiority of our approach for generating authentic and colorful images.
\item A novel plug-and-play \emph{PVCAttn} module is proposed to effectively aggregate color information between source and reference images in an explicable way. 
\item An elaborate \emph{CPM} module is designed to realize continuous magnification, which is more computation-friendly and suitable for practical application.

\end{itemize}

\section{Related Work}
\subsection{Image Colorization}
Before the advent of CNN-based approaches, Li~\etal~\cite{auto_traditional} train a quadratic objective function in the chromaticity maps to colorize images. Subsequently, learning-based approaches almost dominate the automatic image colorization~\cite{survey,auto_dl1,auto_dl2,auto_dl3,auto_dl4}. Cheng~\etal~\cite{auto_dl1} propose to extract multiple-level feature descriptors to regress pixel values, while Zhang~\etal~\cite{auto_dl2} quantize the chrominance space into bins. Later works~\cite{auto_vae1,auto_vae2} leverage VAE to learn a low dimensional embedding of color fields, while GAN-based methods~\cite{auto_gan1,auto_gan2} introduce adversarial training to generate diverse and authentic colorful images. Moreover, I2C~\cite{auto_seg1} uses an off-the-shelf object detector to obtain extra object-level features, while Lei~\etal~\cite{auto_twostage} design a two-stage network successively for colorization and refinement. Even though the above automatic methods perform well, they are uncontrollable and cannot generate various images once trained. This work also considers diversity and controllability when designing the network while retaining the benefits of automatic image colorization.

\begin{figure*}[htb] 
   \centering
   \includegraphics[width=0.90\linewidth]{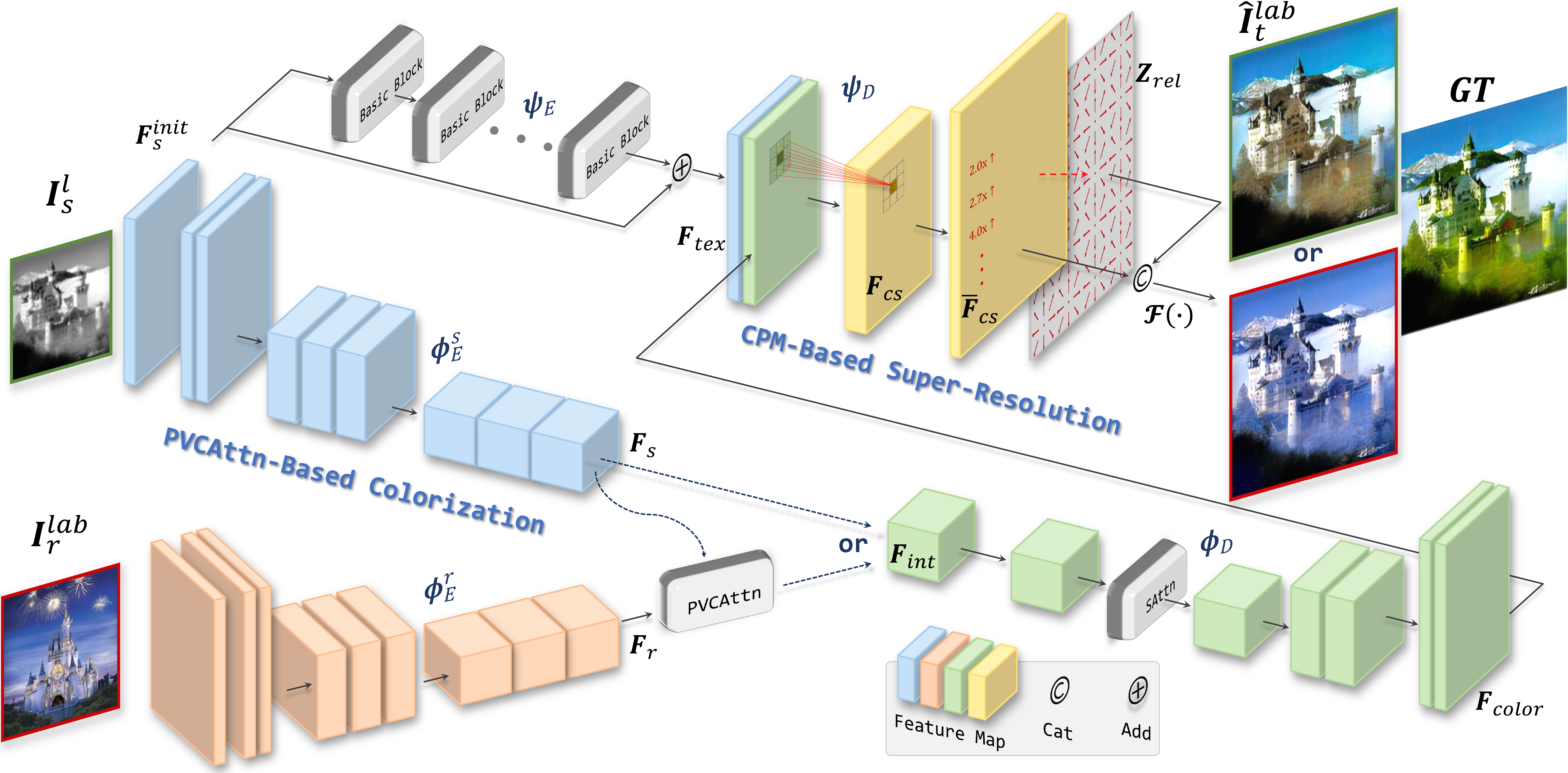}
   \caption{\textbf{Overview of the proposed SCSNet}, which consists of a \emph{PVCAttn-based Colorization} branch ($\boldsymbol{\phi}$) and a \emph{CPM-based Super-resolution} branch ($\boldsymbol{\psi}$). Given a low-resolution gray source image $\boldsymbol{I}_{s}^{l}$ and a colorful reference image $\boldsymbol{I}_{r}^{lab}$, the encoders $\boldsymbol{\phi}_{E}^{s}$ and $\boldsymbol{\phi}_{E}^{r}$ in colorization branch are used to extract corresponding deep features $\boldsymbol{F}_{s}$ and $\boldsymbol{F}_{r}$, respectively. PVCAttn module aggregates $\boldsymbol{F}_{s}$ and $\boldsymbol{F}_{r}$ to form $\boldsymbol{F}_{int}$ if choosing the referential mode, otherwise $\boldsymbol{F}_{int}$ equaling $\boldsymbol{F}_{s}$. Subsequent colorization decoder $\boldsymbol{\phi}_{D}$ restores the color feature to the original resolution $\boldsymbol{F}_{color}$. For super-resolution branch, the encoder $\boldsymbol{\psi}_{E}$ extracts the residual feature of initial feature $\boldsymbol{F}_{s}^{init}$, while the CPM module in decoder $\boldsymbol{\psi}_{D}$ uses mapping function $\boldsymbol{\mathcal{F}}(\cdot)$ to generate target HR image $\boldsymbol{\hat{I}}_{t}^{lab}$ in a continuous space.}
   \label{fig:SCSNet}
\end{figure*}

Differently, referential image colorization requires additional information to guide the generation process. Zou~\etal~\cite{ref_lang3} propose a SECat network that inputs a gray-scale line art and color tag information to produce a quality colored image. Some GAN-based methods~\cite{ref_gan1,ref_gan2,ref_2} use adversarial training to improve the rationality of generated images, while works~\cite{ref_1,ref_3,ref_gan2} take gray sketch image as input and color it with the aid of the reference image condition. He~\etal~\cite{ref_4} propose a similarity sub-net to compute the bidirectional similarity map between source and reference images. Considering the limitation of one-stage network, methods~\cite{ref_twostage1,ref_twostage2} design the coarse-to-fine network to improve the performance. Nevertheless, how to reasonably aggregate the referential feature is still a big challenge~\cite{auto_seg2,ref_twostage1}. Works~\cite{ref_twostage2,ref_1} propose to obtain the correlation matrix whose elements characterize the similarity between the source and reference images.
Recently, Gray2ColorNet~\cite{ref_attn} design an attention gating mechanism-based color fusion network, and Kumar~\etal~\cite{auto_trans} firstly introduce the transformer~\cite{transformer} structure. However, the above methods are effortless to select incorrect referential information prone to produce visual artifacts, \eg, color shift and color patch. To alleviate the problems, we propose a novel PVCAttn module to more effectively aggregate information between source and reference images.

\subsection{Single Image Super-Resolution}
Since Dong~\etal~\cite{sisr_start} propose SRCNN for SISR, many CNN-based methods~\cite{sisr_res1,sisr_res2,sisr_res3,sisr_gan1,sisr_gan2,sisr_loop} with good effects have been proposed. EDSR~\cite{sisr_res1} improves performance significantly by removing unnecessary batch normalization in conventional residual networks and designing a new multi-scale deep super-resolution system. Later RCAN~\cite{sisr_res2} and RDN~\cite{sisr_res3} improve the residual block, and works~\cite{sisr_gan1,sisr_gan2} further introduce adversarial loss during the training phase that greatly improves the model's performance. To solve the problem of real-world image matching, works~\cite{sisr_real1,sisr_real2} contribute new datasets where paired real-world LR-HR images on the same scene are captured. Recently, Guo~\etal~\cite{sisr_loop} propose a novel dual regression scheme for paired and unpaired data, which forms a closed-loop to provide additional supervision. The above methods have achieved good results, but they can only carry out fixed factors for SISR, not producing a continuous display for practical application. Different from Meta-SR that attempts continuous magnification by predicting convolutional weights for each pixel, we design a more efficient \emph{Continuous Pixel Mapping} head to directly regress pixel value with local relative coordinate in a continuous space.


\section{Approach}
In this paper, a novel efficient paradigm is proposed to complete both automatic and referential image colorization along with SISR simultaneously by one end-to-end network. As depicted in Figure~\ref{fig:SCSNet}, the proposed SCSNet consists of a PVCAttn-based colorization branch for restoring the color information, as well as a CPM-based super-resolution branch for generating high-resolution target image in a continuous space. An initial convolution firstly increases the channel dimension of the low-resolution gray-scale source image: $\boldsymbol{I}_{s}^{l} \in\mathbb{R}^{1 \times H_{s} \times W_{s}} \rightarrow \boldsymbol{F}_{s}^{init} \in\mathbb{R}^{64 \times H_{s} \times W_{s}}$, where $H_{s}$ and $W_{s}$ are the height and width of the input image.

\textbf{For the colorization branch}, encoders $\boldsymbol{\phi}_{E}^{s}$ and $\boldsymbol{\phi}_{E}^{r}$ are employed to extract corresponding deep features:

\begin{equation}
   \begin{aligned}
      \boldsymbol{F}_{s} = &\boldsymbol{\phi}_{E}^{s}(\boldsymbol{F}_{s}^{init}) \in\mathbb{R}^{256 \times H_{s}/4 \times W_{s}/4} \text{,} \\
      \boldsymbol{F}_{r} = &\boldsymbol{\phi}_{E}^{r}(\boldsymbol{I}_{r}^{lab}) \in\mathbb{R}^{256 \times H_{s}/4 \times W_{s}/4} \text{.}
   \end{aligned}
\end{equation}
We design the branch with two patterns: the automatic mode that directly maps source image feature to output (\ie, $\boldsymbol{F}_{s} \rightarrow \boldsymbol{F}_{int}$), and the referential mode that employs the proposed plug-and-play PVCAttn module to aggregate both source and reference image features:

\begin{equation}
   \begin{aligned}
      \boldsymbol{F}_{int} = \text{PVCAttn}(\boldsymbol{F}_{s}, \boldsymbol{F}_{r}) \in\mathbb{R}^{256 \times H_{s}/4 \times W_{s}/4} \text{.}
   \end{aligned}
\end{equation}
Subsequently, decoder $\boldsymbol{\phi}_{D}$ restores the color information $\boldsymbol{F}_{int}$ to the original resolution $\boldsymbol{F}_{color} \in\mathbb{R}^{64 \times H_{s} \times W_{s}}$ via a self-attention layer and several convolution layers.

\textbf{For super-resolution branch}, encoder $\boldsymbol{\psi}_{E}$ extracts the residual texture feature $\boldsymbol{F}_{tex} \in\mathbb{R}^{64 \times H_{s} \times W_{s}}$ from the initial feature map $\boldsymbol{F}_{s}^{init}$ via concatenated basic blocks. Each basic block contains two convolution layers along with a skip operation. A subsequent 3$\times$3 convolution is used to aggregate $\boldsymbol{F}_{tex}$ and $\boldsymbol{F}_{color}$, indicated as $\boldsymbol{F}_{cs} \in\mathbb{R}^{256 \times H_{s} \times W_{s}}$. Finally, the CPM module employs mapping function $\boldsymbol{\mathcal{F}}(\cdot)$ to regress target HR image $\boldsymbol{\hat{I}}_{t}^{lab} \in\mathbb{R}^{3 \times H_{s}*p \times W_{s}*p}$, and $p$ represents any magnification that can be a decimal, while $\boldsymbol{\overline{F}}_{cs}\in\mathbb{R}^{2 \times H_{s}*p \times W_{s}*p}$ is obtained according to $\boldsymbol{F}_{cs}$.

\begin{figure}[t]
   \centering
   \includegraphics[width=0.90\columnwidth]{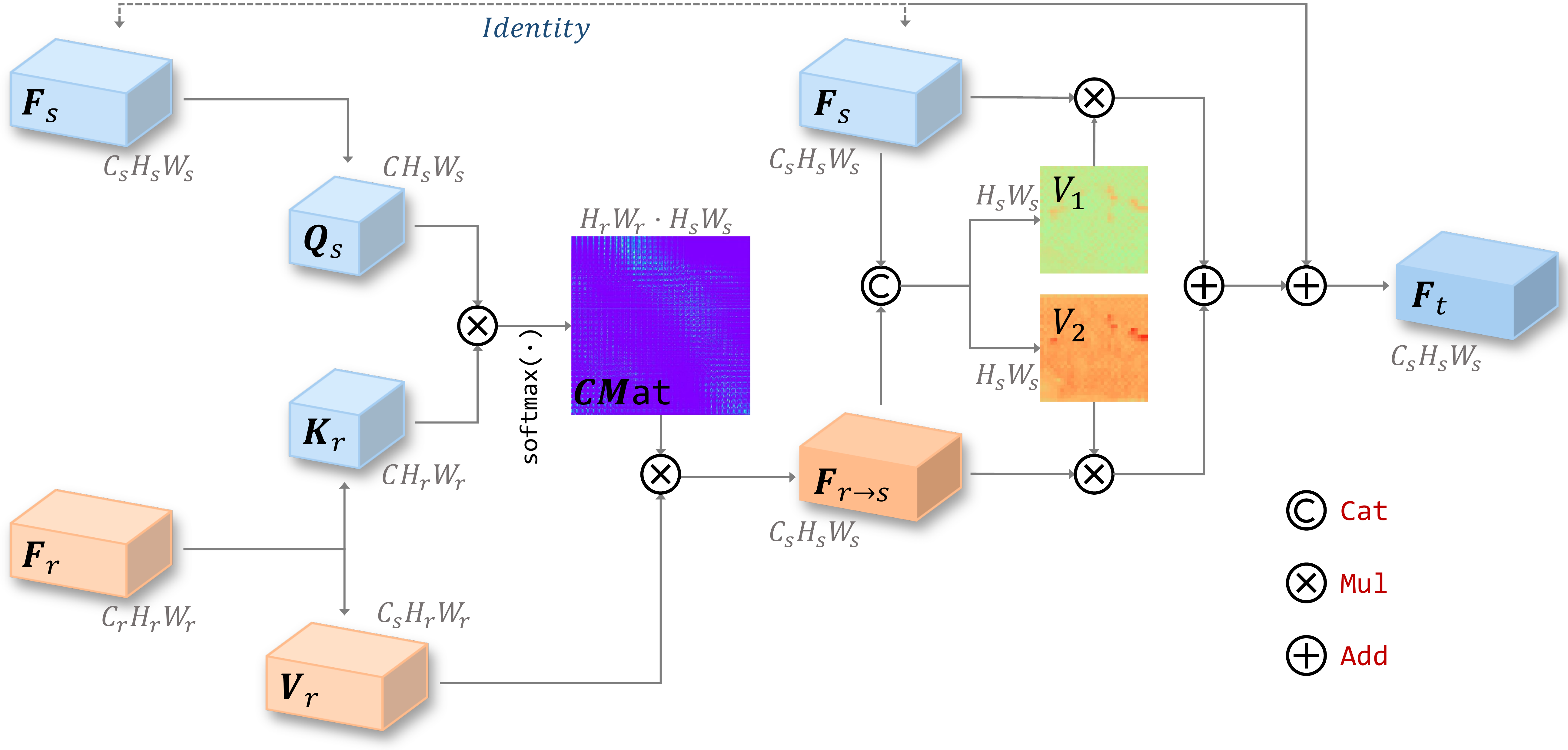}
   \caption{\textbf{Schematic diagram of VCAttn}. The module takes the source feature $\boldsymbol{F}_{s}$ and the reference feature $\boldsymbol{F}_{r}$ as input and outputs the aggregated target feature map $\boldsymbol{F}_{t}$ that has the same dimension with $\boldsymbol{F}_{s}$. \emph{Identity} means the shared feature map.}
   \label{fig:VCAttn}
\end{figure}

\subsection{Pyramid Valve Cross Attention}
\begin{figure}[t]
   \centering
   \includegraphics[width=0.90\columnwidth]{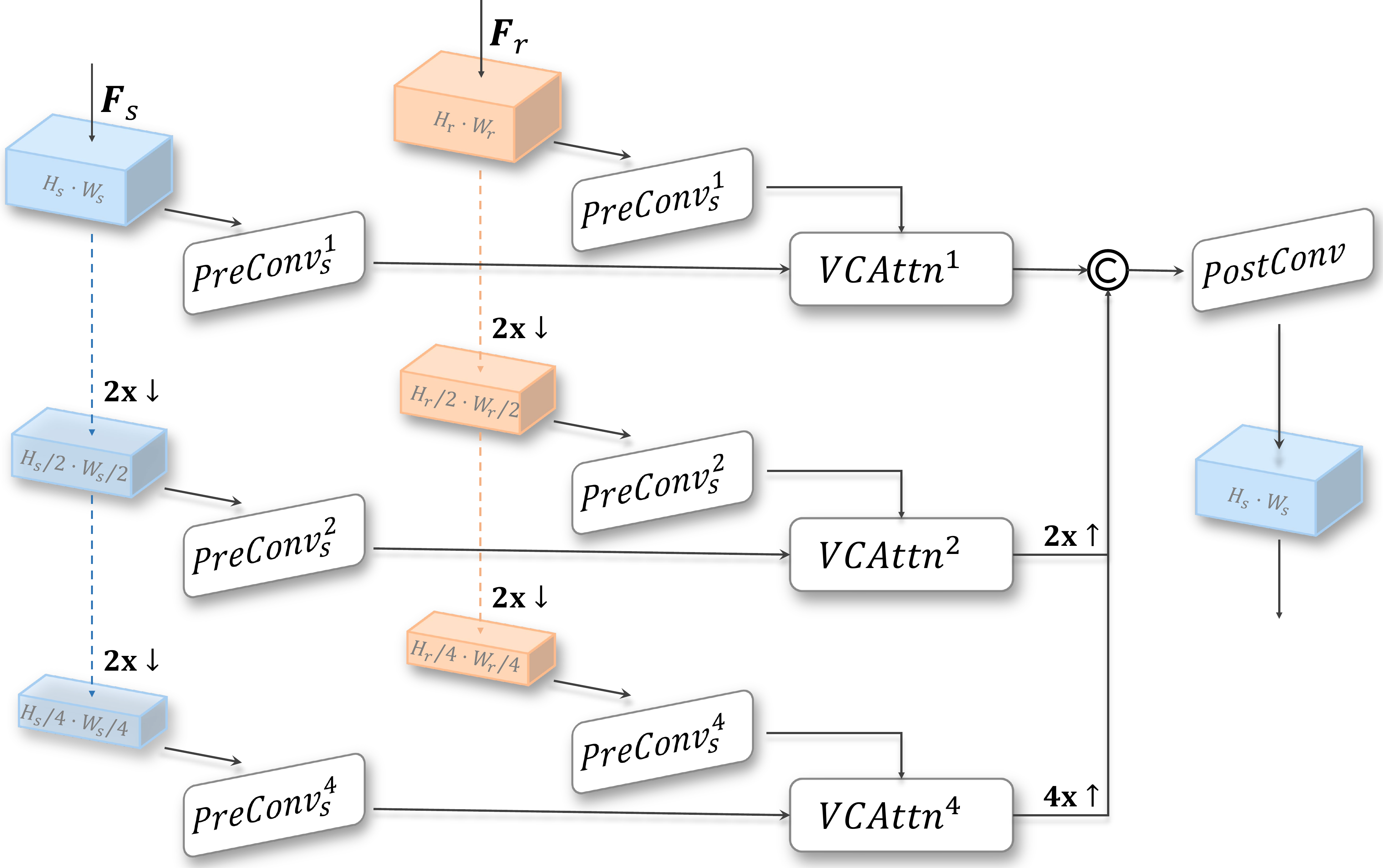}
   \caption{\textbf{Schematic diagram of PVCAttn}. Pre-convolved pyramid feature maps are processed by multiple VCAttn modules.}
   \label{fig:PVCAttn}
\end{figure}

In order to more effectively aggregate feature information between source and reference images, we propose a novel \emph{Valve Cross Attention} (VCAttn) module. As illustrated in Figure~\ref{fig:VCAttn}, the purpose of VCAttn is to select the reference feature $\boldsymbol{F}_{r} \in\mathbb{R}^{C_{r} \times H_{r} \times W_{r}}$ reasonably to the source feature $\boldsymbol{F}_{s} \in\mathbb{R}^{C_{s} \times H_{s} \times W_{s}}$. Similar to SAttn~\cite{sagan}, three convolution operations are used to extract query features $\boldsymbol{Q}_{s} \in\mathbb{R}^{C \times H_{s} \times W_{s}}$, key features $\boldsymbol{K}_{r} \in\mathbb{R}^{C \times H_{r} \times W_{r}}$, and value features $\boldsymbol{V}_{r} \in\mathbb{R}^{C_{s} \times H_{r} \times W_{r}}$, respectively. Then, $\boldsymbol{Q}_{s}$ and $\boldsymbol{K}_{r}$ are employed to calculate the correlation matrix $\boldsymbol{CMat}$, which further multiplies $\boldsymbol{V}_{r}$ to obtain $\boldsymbol{F}_{r \rightarrow s}$. Subsequently, concatenated $\boldsymbol{F}_{s}$ and $\boldsymbol{F}_{r \rightarrow s}$ go through cascaded 1$\times$1 Convolution and Sigmoid to obtain valve maps $\boldsymbol{V}_{1}$ and $\boldsymbol{V}_{2}$, which are used to control the information flux of $\boldsymbol{F}_{s}$ and $\boldsymbol{F}_{r \rightarrow s}$. To further improve the representation, we design a pyramid VCAttn module (PVCAttn) in Figure~\ref{fig:PVCAttn}: pyramid feature maps are sent into corresponding VCAttn modules after pre-convolving, and the concatenated feature map goes through a post-convolution to obtain the final output.

\begin{figure}[t]
   \centering
   \includegraphics[width=0.90\columnwidth]{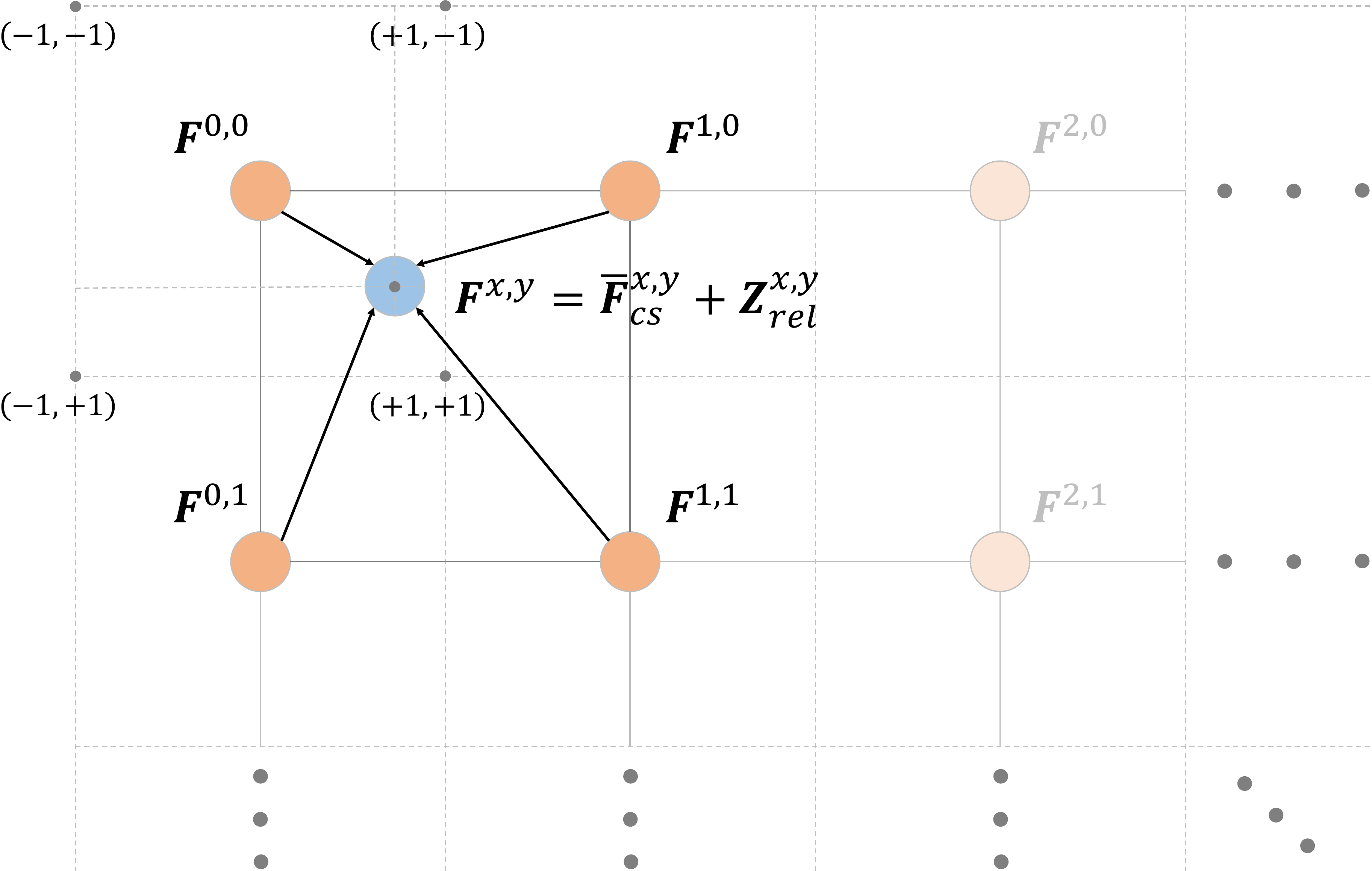}
   \caption{\textbf{Schematic diagram of CPM}. Taking four points on the feature map as an example, the feature $\boldsymbol{F}^{x,y}$ of the point $x,y$ is modeled as the fusion of two parts: one is the main feature $\boldsymbol{\overline{F}}_{cs}^{x,y}$ obtained by bilinear interpolation with corner alignment, the other is the local relative coordinate $\boldsymbol{Z}_{rel}^{x,y}$ to the nearest anchor point.}
   \label{fig:CPM}
\end{figure}

\subsection{Continuous Pixel Mapping} \label{sec:CPM}
In order to generate target images at any magnification, we model the discrete feature mapping in continuous pixel space and propose an efficient super-resolution head named \emph{Continuous Pixel Mapping}. As shown in Figure~\ref{fig:CPM}, we model the feature $\boldsymbol{F}^{x,y}$ of each point $x,y$ with two parts: main feature $\boldsymbol{\overline{F}}_{cs}^{x,y}$ obtained by bilinear interpolation around the neighborhood four points, as well as the local coordinate feature $\boldsymbol{Z}_{rel}^{x,y}$ that describes continuous local spatial information. We model each point in the target image by its local relative coordinate to the nearest point in the original resolution image (\ie, anchor point) for providing fine-grained guidance for each location. Since the coordinate is continuous that it can be infinitely interpolated, and is independent of the image resolution, $\boldsymbol{F}^{x,y}$ can be modeled in a continuous space. Note that we align the corner when obtaining the main feature of each point. As for local coordinate feature $\boldsymbol{Z}_{rel}^{x,y}$, we look for its corresponding anchor point in the original feature $\boldsymbol{F}_{cs} \in\mathbb{R}^{256 \times H_{s} \times W_{s}}$ and calculate local coordinate feature $\boldsymbol{Z}_{rel}^{x,y}$ in the following formula:

\begin{equation}
   \begin{aligned}
      \boldsymbol{Z}_{rel}^{x} = & \text{mod} (x, x_{unit}) / x_{unit} * 2 - 1 \text{,} \\
      \boldsymbol{Z}_{rel}^{y} = & \text{mod} (y, y_{unit}) / y_{unit} * 2 - 1 \text{,}
   \end{aligned}
\end{equation}
where $x_{unit} = 1 / W_{s}, y_{unit} = 1 / H_{s}$, $\text{mod}$ is remainder operation, and $\boldsymbol{Z}_{rel}^{x/y}$ are in range -1 to +1, \ie, (-1,-1) for the upper left corner while (+1,+1) for the lower right corner. Finally, continuous pix mapping function $\boldsymbol{\mathcal{F}}(\cdot)$ that contains four linear layers maps the feature to target image $\boldsymbol{\hat{I}}_{t}^{lab}$.

\subsection{Objective Functions}
During the training stage of SCSNet, we only adopt three losses: \emph{Content Loss} $\mathcal{L}_{C}$ to monitor image quality at the pixel level, \emph{Perceptual Loss} $\mathcal{L}_{P}$ to ensure semantic similarity, and \emph{Adversarial Loss} $\mathcal{L}_{Adv}$ to improve image quality and authenticity. The full loss $\mathcal{L}_{all}$ is defined as follow:
\begin{align}
   \mathcal{L}_{all} = \lambda_{C} \mathcal{L}_{C} + \lambda_{P} \mathcal{L}_{P} + \lambda_{Adv} \mathcal{L}_{Adv},
\end{align}
where $\lambda_{C}=10$, $\lambda_{P}=5$, and $\lambda_{Adv}=1$ represent weight parameters to balance different terms.

\noindent\textbf{Content Loss.} The first term $\mathcal{L}_{C}$ calculates the $\ell_1$ error between the generated target image $\boldsymbol{\hat{I}}_{t}^{lab}$ and ground truth $\boldsymbol{I}_{t}^{lab}$:

\begin{align}
   \mathcal{L}_{C} &= || \boldsymbol{\hat{I}}_{t}^{lab} - \boldsymbol{I}_{t}^{lab} ||_{1} \text{.}
   \label{eq:content}
\end{align}

\noindent\textbf{Perceptual Loss.} The second term $\mathcal{L}_{P}$ calculates semantic errors between the generated target image $\boldsymbol{\hat{I}}_{t}^{lab}$ and the ground truth image $\boldsymbol{I}_{t}^{lab}$:

\begin{align}
   \mathcal{L}_{P} &= \mathbb{E} \left[\sum_{l=1}^{5} w_{l} \cdot || \phi_{l}(\boldsymbol{\hat{I}}_{t}^{lab}) - \phi_{l}(\boldsymbol{I}_{t}^{lab}) ||_{1} \right] \text{,}
   \label{eq:perceptual}
\end{align}
where $\phi_{l}(\cdot)$ represents the activation map extracted at the $convl\_1$ layer from the pre-trained VGG16 network, and $w_{l}$ is the weight for layer$\_l$.

\noindent\textbf{Adversarial Loss.} The third term $\mathcal{L}_{Adv}$ employs the standard relativistic discriminator~\cite{relativisticgan} for adversarial training in order to ensure the authenticity of the generated images. Since the SCS task is typically a one-to-many problem, the adversarial loss greatly improves the model performance.

\begin{equation}
   \begin{aligned}
      \mathcal{L}_{Adv}^{G} = & \mathbb{E}_{{x} \sim p_{x}} \left[(D(x) - \mathbb{E}_{{\tilde{x}} \sim p_{\tilde{x}}}[D(\tilde{x})] - 1)^{2} \right] + \\
                              & \mathbb{E}_{{\tilde{x}} \sim p_{\tilde{x}}} \left[(D(\tilde{x}) - \mathbb{E}_{{x} \sim p_{x}}[D(x)])^{2} \right] \text{,} \\
      \mathcal{L}_{Adv}^{D} = & \mathbb{E}_{{\tilde{x}} \sim p_{\tilde{x}}} \left[(D(\tilde{x}) - \mathbb{E}_{{x} \sim p_{x}}[D(x)] - 1)^{2} \right] \text{,}
      \label{eq:adversarial}
   \end{aligned}
\end{equation}
where $p_{x}$ and $p_{\tilde{x}}$ are real and generated image distributions.

\begin{figure*}[htb] 
   \centering
   \includegraphics[width=0.9\linewidth]{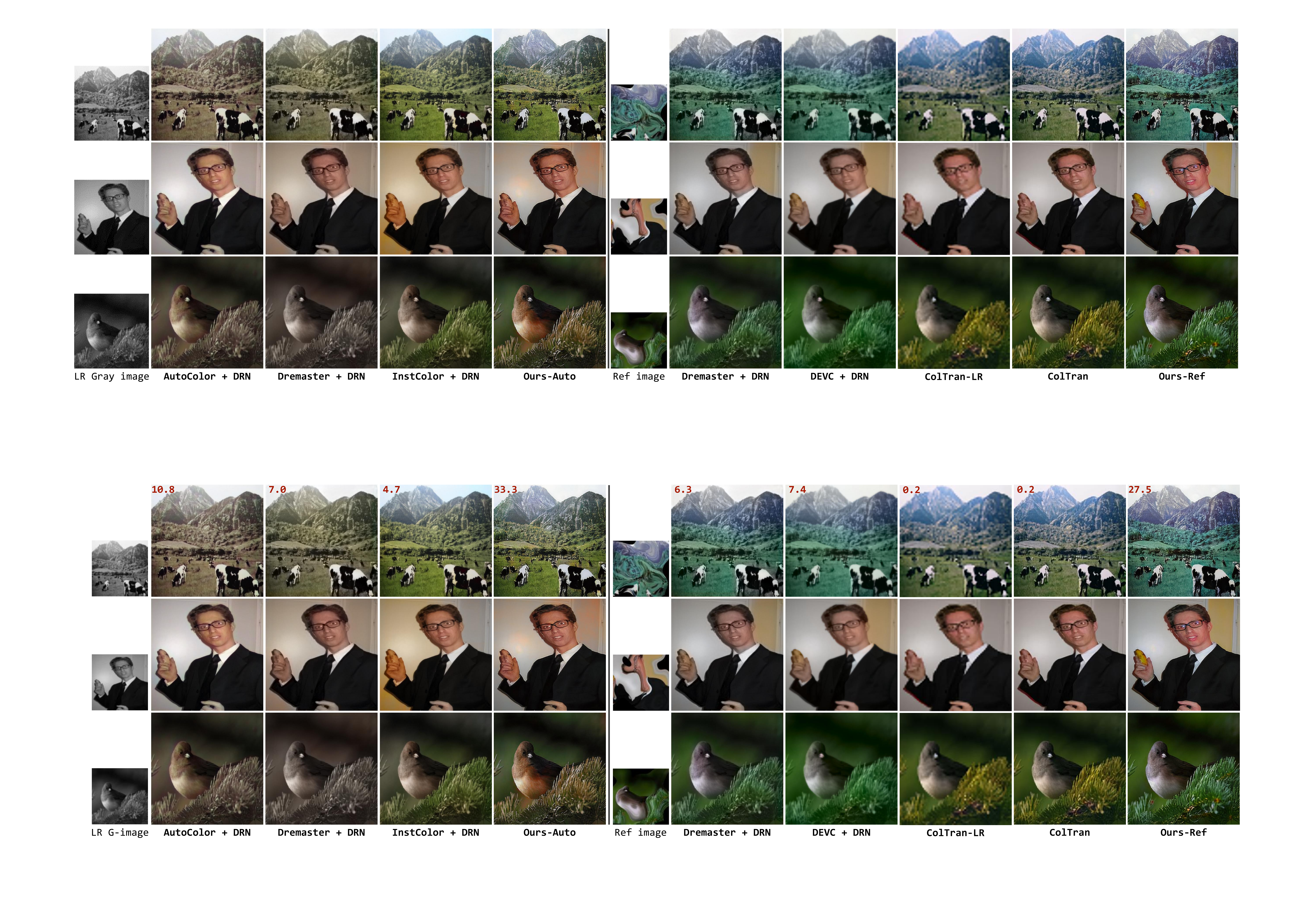}
   \caption{\textbf{Qualitative comparisons with SOTA methods on ImageNet-C and COCO datasets.} The left and right parts are \emph{simultaneously image colorization and super-resolution} results in automatic and referential modes, respectively. The upper left number is the inference FPS speed of the corresponding method. Note that ColTran inputs an \emph{extra} HR gray-scale image that is unfair for other approaches, so we reduce its input gray-scale image to the same resolution as other methods, \ie, ColTran-LR.}
   \label{fig:Coms}
\end{figure*}

\section{Experiments}
\subsection{Datasets and Implementation Details}
\paragraph{ImageNet-C.} Considering the high requirement for both image colorization and super-resolution, we filtered out some of the low-quality images from ImageNet~\cite{imagenet} to build a colorful and high-resolution dataset named ImageNet-C. It ends up with 407,041 training images and 16,216 validation images. In detail, we remove images with file sizes smaller than 80K and less color variation.

\paragraph{Other Datasets.} CelebA-HQ~\cite{celebahq} (30,000 images), Flowers~\cite{102flowers} (8,189 images), Bird~\cite{lsun} (479,548 images), and COCO~\cite{coco} (98,246 images) datasets are used to assess different colorization methods (for testing), and they go through the same pre-selection process as ImageNet-C.

\paragraph{Evaluation Metrics.} We use \emph{Peak Signal-to-Noise Ratio} (PSNR) and \emph{Structural Similarity} (SSIM)~\cite{ssim} to assess the generated images at pixel level, while \emph{Fr\'echet Inception Distance} (FID)~\cite{fid} and \emph{Image ColorfulNess} (CN)~\cite{colornet} to assess image distribution and colorfulness at semantic level. In order to fully evaluate various methods, we hire real people to score authenticity of images for human study.

\paragraph{Implementation Details.} The image is processed in LAB color space, and the input resolution of source and reference images is 128$\times$128. Consistent with DRN, the bicubic kernel is used to produce LR-HR pairs. We train the SCSNet with two modes alternately and apply random horizontal flip and elastic distortion~\cite{elastic2} to reference images. Perceptual weights $w_{1-5}$ in Eq.~\ref{eq:perceptual} are set as 1.0/32, 1.0/16, 1.0/8, 1.0/4, and 1.0, respectively. We use Adam~\cite{adam} optimizer and set $\beta_1=0.9$, $\beta_2=0.999$, weight-decay=$1e^{-4}$, and learning rate=$1e^{-4}$. SCSNet is trained for 50 epochs with batch-size$=$4 and output-resolution$=$512 (default $\times$4$\uparrow$ setting). Resolution of the referential image for all experiments is set to 128, and all experiments run with 8 Tesla V100 GPUs.

\begin{table*}[t]
   \centering
   \scriptsize
   \renewcommand\arraystretch{0.8}
   \setlength\tabcolsep{5pt}
   \begin{tabular}{C{5pt}C{68pt}C{17pt}C{17pt}C{17pt}C{17pt}C{17pt}C{17pt}C{17pt}C{17pt}C{17pt}C{17pt}C{17pt}C{17pt}C{17pt}C{17pt}}
      \toprule
      & \multicolumn{1}{c}{\multirow{2}{*}{Method}} & \multicolumn{2}{c}{ImageNet-C} & \multicolumn{2}{c}{CelebA-HQ} & \multicolumn{2}{c}{Flowers} & \multicolumn{2}{c}{Bird} & \multicolumn{2}{c}{COCO} & \multicolumn{2}{c}{Average} & \multicolumn{1}{c}{\multirow{2}{*}{\makecell[c]{Params\\(M)$\downarrow$}}} & \multicolumn{1}{c}{\multirow{2}{*}{\makecell[c]{Speed\\(FPS)$\uparrow$}}} \\
      \cmidrule(lr){3-4} \cmidrule(lr){5-6} \cmidrule(lr){7-8} \cmidrule(lr){9-10} \cmidrule(lr){11-12} \cmidrule(lr){13-14}
      & & FID $\downarrow$ & CN $\uparrow$ & FID $\downarrow$ & CN $\uparrow$ & FID $\downarrow$ & CN $\uparrow$ & FID $\downarrow$ & CN $\uparrow$ & FID $\downarrow$ & CN $\uparrow$ & FID $\downarrow$ & CN $\uparrow$ & & \\
      \midrule
      & Real Test Dataset & - & 5.522 & - & 4.344 & - & 5.783 & - & 5.006 & - & 5.222 & - & 5.175 & - & - \\
      \midrule
	   \multirow{7}{*}{{\rotatebox[origin=c]{90}{Automatic}}}
      & AutoColor~~+~ESRGAN      & 31.725 & 3.390 & 25.637 & 3.488 & 64.107 & 3.210 & 42.956 & 3.190 & 38.904 & 3.417 & 40.666 & 3.339 & 25.432 & 16.910 \\
      & DRemaster~+~ESRGAN       & 33.908 & 2.998 & 34.775 & 3.054 & 74.601 & 2.678 & 45.505 & 2.833 & 37.249 & 3.078 & 45.208 & 2.928 & 65.516 & 9.217 \\
      & ~InstColor~~~+~ESRGAN    & 26.353 & 3.588 & 31.109 & 3.306 & 48.425 & 3.510 & 36.212 & 3.435 & 24.954 & 3.635 & 33.411 & 3.495 & 66.990 & 5.541 \\
      \cmidrule(lr){2-16}
      & AutoColor~~+~DRN         & 31.666 & 3.389 & \textbf{26.083} & \underline{3.489} & 63.891 & 3.210 & 43.034 & 3.189 & 39.208 & 3.417 & 40.776 & 3.339 & \underline{18.560} & \underline{10.781} \\
      & DRemaster~+~DRN          & 33.993 & 2.996 & 34.842 & 3.054 & 74.356 & 2.678 & 45.742 & 2.832 & 37.397 & 3.076 & 45.266 & 2.927 & 58.644 & 7.037 \\
      & ~InstColor~~~+~DRN       & \underline{26.501} & \underline{3.588} & 31.389 & 3.307 & \underline{48.475} & \underline{3.511} & \underline{36.378} & \underline{3.436} & \underline{25.153} & \underline{3.635} & \underline{33.579} & \underline{3.495} & 60.118 & 4.671 \\
      & Ours-Auto                & \textbf{25.992} & \textbf{4.688} & \underline{27.809} & \textbf{3.892} & \textbf{46.607} & \textbf{4.724} & \textbf{34.401} & \textbf{4.334} & \textbf{24.047} & \textbf{4.573} & \textbf{31.771} & \textbf{4.442} & \textbf{9.954}  & \textbf{33.293} \\
      \midrule
	   \multirow{4}{*}{{\rotatebox[origin=c]{90}{Referential}}}
      & DRemaster~+~DRN          & 25.498 & 3.990 & 29.702 & 3.426 & 41.506 & 4.311 & 31.843 & 3.505 & 35.267 & 3.722 & 32.763 & 3.791 & 73.987 & 6.311 \\
      & ~~~~~~~DEVC~+~DRN        & 26.050 & 4.288 & 49.126 & 3.570 & 39.426 & 4.516 & 42.444 & 3.700 & 36.702 & 3.938 & 38.750 & 4.002 & \underline{69.570} & \underline{7.435} \\
      & ColTran                  & \underline{15.860} & \underline{4.692} & \textbf{10.405} & \underline{4.215} & \underline{21.595} & \underline{5.003} & \underline{18.580} & \underline{4.529} & \underline{18.391}& \underline{4.679} & \underline{16.966} & \underline{4.624} & 70.697 & 0.209 \\
      & Ours-Ref                 & \textbf{9.632} & \textbf{5.288} & \underline{12.771} & \textbf{4.388} & \textbf{9.776} & \textbf{5.743} & \textbf{13.526} & \textbf{4.822} & \textbf{13.812} & \textbf{4.974} & \textbf{11.903} & \textbf{5.043} & \textbf{15.358} & \textbf{27.466} \\
      \bottomrule
   \end{tabular}
   \caption{\textbf{Image-level evaluation for SOTA methods on several datasets.} Since the SCS task is an ill-conditioned problem that each pixel has various semantic colors, more reasonable image-level FID and CN are used. \textbf{Bold} and \underline{underline} represent optimal and suboptimal results.}
   \label{tab:comparison}
\end{table*}

\begin{table}[t]
   \centering
   \scriptsize
   \renewcommand\arraystretch{0.8}
   \setlength\tabcolsep{2pt}
   \begin{tabular}{C{55pt}C{25pt}C{25pt}C{25pt}C{25pt}C{25pt}C{25pt}}
      \toprule
      \multicolumn{1}{c}{\multirow{2}{*}{Method}} & \multicolumn{2}{c}{ImageNet-C} & \multicolumn{2}{c}{CelebA-HQ} & \multicolumn{2}{c}{COCO} \\
      \cmidrule(lr){2-3} \cmidrule(lr){4-5} \cmidrule(lr){6-7} 
      & PSNR $\uparrow$ & SSIM $\uparrow$ & PSNR $\uparrow$ & SSIM $\uparrow$ & PSNR $\uparrow$ & SSIM $\uparrow$ \\
      \midrule
      DRemaster~+~DRN   & 19.343 & 0.811 & 25.559 & 0.915 & 21.039 & 0.845 \\
      ~InstColor~~~+~DRN& \underline{22.126} & \underline{0.842} & \underline{26.523} & \textbf{0.923} & \underline{22.917} & \underline{0.856} \\
      Ours-Auto         & \textbf{22.807} & \textbf{0.856} & \textbf{27.160} & \underline{0.917} & \textbf{23.341} & \textbf{0.872} \\
      \midrule
      ColTran           & 20.734 & 0.845 & 24.495 & 0.914 & 22.787 & 0.857 \\
      DRemaster~+~DRN   & \underline{24.671} & \underline{0.871} & \underline{28.582} & \underline{0.928} & \underline{26.663} & \underline{0.901} \\
      Ours-Ref          & \textbf{27.694} & \textbf{0.923} & \textbf{30.741} & \textbf{0.950} & \textbf{28.197} & \textbf{0.931} \\
      \bottomrule
   \end{tabular}
   \caption{\textbf{Pixel-level evaluation for SOTA methods.} Top and bottom parts are for automatic and referential modes, respectively.}
   \label{tab:comparison1}
\end{table}

\subsection{Comparison with SOTAs}
We conduct and discuss a series of qualitative and quantitative comparison experiments on several datasets. At present, there is no end-to-end model to perform image colorization and super-resolution simultaneously, so we choose some SOTA colorization methods (\ie, AutoColor~\cite{auto_twostage}, DRemaster~\cite{ref_4}, InstColor~\cite{auto_seg1}, DEVC~\cite{ref_twostage2}, ColTran~\cite{auto_trans}) along with concatenated super-resolution approaches (ESRGAN~\cite{sisr_gan2}, DRN~\cite{sisr_loop}) as our comparison methods. Concretely, we divide the above colorization methods into automatic and referential modes.

\noindent\textbf{Qualitative Results.}
We conduct a series of qualitative experiments on ImageNet-C and COCO validation datasets to visually show the superiority of our approach for generating authentic and colorful images for the SCS problem. As shown in Figure~\ref{fig:Coms}, the left part shows automatic SCS results of different methods that use the low-resolution gray-scale images (first column) as input. All methods can distinguish semantic targets and color them, but our generated images look better in colorfulness and detail than other approaches. The right part shows results for various methods under the condition of an elastic reference image in resolution 128$\times$128, which provides the color information that the real image should contain. All methods could transfer referential color well except ColTran, but our method can produce clearer and authentic images while maintaining color transfer. Note that ColTran inputs an extra HR gray-scale image for better clarity that is unfair for the SCS task. We reduce its input gray-scale image to the same resolution as other methods, and the output images get a little blurry (\cf ColTran-LR in the penultimate column).

\noindent\textbf{Quantitative Results.}
We choose image-level metrics to evaluate the effectiveness of different SOTA methods on several datasets: FID for assessing image distribution while visual-friendly CN for colorfulness. Our approach is trained only on ImageNet-C without extra datasets, while other methods use corresponding pre-trained models that may use extra datasets for training. We randomly choose 5,000 images of each method for assessment (2,500 under automatic mode; 500 by self-referential elastic images; while 2,000 by randomly selecting other elastic images as reference images). Table~\ref{tab:comparison} shows the results of different methods for several datasets on two modes, and we can summarize the following conclusions: \textbf{1)} The middle part illustrates that different SR methods have little difference in the results, so we choose SOTA DRN for SISR in the following experiments. \textbf{2)} Referential mode tends to get better results than automatic mode. \textbf{3)} Different datasets are slightly different in CN metric and our method obtains the highest CN score (\ie, 4.442 and 5.043 for two modes, increasing +0.947 and +0.419 than current best results), meaning that our approach can capture color information better and generate visual-appealing colorful images. \textbf{4)} Our method obtains the best FID scores on almost all datasets no matter in automatic mode or referential mode, meaning that the generated images by our method have a more consistent distribution of the real images. \textbf{5)} We further evaluate the parameter and running speed of all approaches, and our SCSNet has the fewest parameters ($\times$6$\downarrow$ than InstColor+DRN; $\times$4$\downarrow$ than ColTran) and fastest running speed ($\times$8$\uparrow$ than InstColor+DRN; $\times$130$\uparrow$ than ColTran; with batch size equaling one), which is more efficient for practical application. Furthermore, we use pixel-level PSNR and SSIM to evaluate generated images under automatic and self-referential modes. As shown in Table~\ref{tab:comparison1}, our method consistently obtains better evaluation scores, meaning that the predicted images by SCSNet are more consistent with real images. Interestingly, the aforementioned ColTran inability to integrate referential image colors (\cf Figure~\ref{fig:Coms}) is also reflected here, where it obtains worse pixel-level metric scores.

\begin{table}[t]
   \centering
   \scriptsize
   \renewcommand\arraystretch{1.0}
   \setlength\tabcolsep{6pt}
   \begin{tabular}{C{90pt}C{60pt}}
      \toprule
      Comparison Methods & Authenticity (\%) \\
      \midrule
      Ours \textit{v.s.} AutoColor~~+~DRN & \textbf{77.3} \textit{v.s.} 22.7 \\
      Ours \textit{v.s.} DRemaster~+~DRN & \textbf{90.6} \textit{v.s.} 9.4~~~  \\
      Ours \textit{v.s.} InstColor~~~~+~DRN~ & \textbf{59.4} \textit{v.s.} 40.6 \\
      \midrule
      Ours \textit{v.s.} DRemaster~+~DRN  & \textbf{82.7} \textit{v.s.} 17.3 \\
      Ours \textit{v.s.} DEVC~~~~~~~~+~DRN~       & \textbf{95.1} \textit{v.s.} 4.9~~~  \\
      Ours \textit{v.s.} ColTran~~~~~~~~~~~~~~~~~~~    & \textbf{68.6} \textit{v.s.} 31.4 \\
      \bottomrule
   \end{tabular}
   \caption{Human study about the authenticity of generated images with different methods in automatic and referential modes.}
   \label{tab:amt}
\end{table}

\begin{table}[t]
   \centering
   \scriptsize
   \renewcommand\arraystretch{1.0}
   \setlength\tabcolsep{5pt}
   \begin{tabular}{C{20pt}C{20pt}C{20pt}C{25pt}C{25pt}C{25pt}C{25pt}}
      \toprule
      $\mathcal{L}_{C}$ & $\mathcal{L}_{P}$ & $\mathcal{L}_{Adv}$ & FID $\downarrow$ & CN $\uparrow$ & PSNR $\uparrow$ & SSIM $\uparrow$ \\
      \midrule
      \cmark & \xmarkg  & \xmarkg   & 16.290	            & 4.128              & 26.172             & 0.898              \\
      \cmark & \cmark   & \xmarkg   & 15.068	            & 4.196              & \underline{27.456} & \underline{0.918}  \\
      \cmark & \xmarkg  & \cmark    & \underline{11.616}	& \underline{5.171}  & 26.536             & 0.906              \\
      \cmark & \cmark   & \cmark    & \textbf{9.632}	   & \textbf{5.288}     & \textbf{27.694}    & \textbf{0.923}     \\
      \bottomrule
   \end{tabular}
   \caption{Quantitative ablation study for different loss terms.}
   \label{tab:ablation_loss}
\end{table}

\begin{table}[t]
   \centering
   \scriptsize
   \renewcommand\arraystretch{1.0}
   \setlength\tabcolsep{4pt}
   \begin{tabular}{C{18pt}C{18pt}C{18pt}C{18pt}C{25pt}C{25pt}C{25pt}C{25pt}}
      \toprule
      Baseline & BCAttn & PVCAttn & CPM & FID $\downarrow$ & CN $\uparrow$ & PSNR $\uparrow$ & SSIM $\uparrow$ \\
      \midrule
      \cmark & \xmarkg  & \xmarkg   & \xmarkg   & 17.541	            & 4.763              & 25.517             & 0.887              \\
      \cmark & \cmark   & \xmarkg   & \xmarkg   & 16.635	            & 4.767              & 26.173             & 0.896              \\
      \cmark & \xmarkg  & \cmark    & \xmarkg   & 15.334	            & 4.863              & 26.619             & \underline{0.907}  \\
      \cmark & \xmarkg  & \xmarkg   & \cmark    & \underline{9.978}	& \underline{5.059}  & \underline{26.796} & 0.905              \\
      \cmark & \xmarkg  & \cmark    & \cmark    & \textbf{9.632}	   & \textbf{5.288}     & \textbf{27.694}    & \textbf{0.923}     \\
      \bottomrule
   \end{tabular}
   \caption{Quantitative ablation study of our approach with different components on the ImageNet-C dataset.}
   \label{tab:ablation_module}
\end{table}

\noindent\textbf{Human Study.} 
Since SCS is an ill-conditioned problem, and each metric has its evaluation disadvantage, we further perform a human study for artificially evaluating the quality of generated images for different methods. Concretely, we randomly select 500 generated images (250:250 for two modes) of different approaches on the COCO dataset. Each image pair (\ie, ours \vs each other method) is displayed for one second for each conner (50 totally), and the conner needs to select which image is more visually authentic. Table~\ref{tab:amt} illustrates that the generated images by our approach are preferred by real people, meaning that our method can generate more authentic images than SOTA methods. 

\subsection{Ablation Study and Further Assessment}

\noindent\textbf{Loss Functions.} Following the afore-mentioned procedure for generating the validation images (under referential mode), we quantitatively evaluate the effectiveness of each loss function in Table~\ref{tab:ablation_loss} and draw a conclusion: Each loss function contributes to the model performance, and the model obtains the best score when all loss terms are applied.

\noindent\textbf{Network Components.} We perform quantitative experiments to evaluate each component of our approach. Specifically, we modify a simple version of PVCAttn as \emph{Basic Cross-Attention} (BCAttn) that removes pyramid structure and valves, which is used for a fair comparison with our PVCAttn. Results in Table~\ref{tab:ablation_module} demonstrate the effectiveness of each component, and our approach obtains the highest metric scores when both proposed components are used. Moreover, the CPM module obtains competitive results even though it is designed for continuous magnification.

\noindent\textbf{CPM Efficiency.} We compare CPM module with Meta-SR~\cite{metasr} that also achieves continuous magnification, and results indicate that CPM is more efficient as it has fewer parameters and a $\times$2$\uparrow$ faster running speed, \ie, \textbf{0.35M} \vs 0.45M and \textbf{178FPS} \vs 92 FPS.

\begin{figure}[!tp]
   \centering
   \includegraphics[width=1.0\columnwidth]{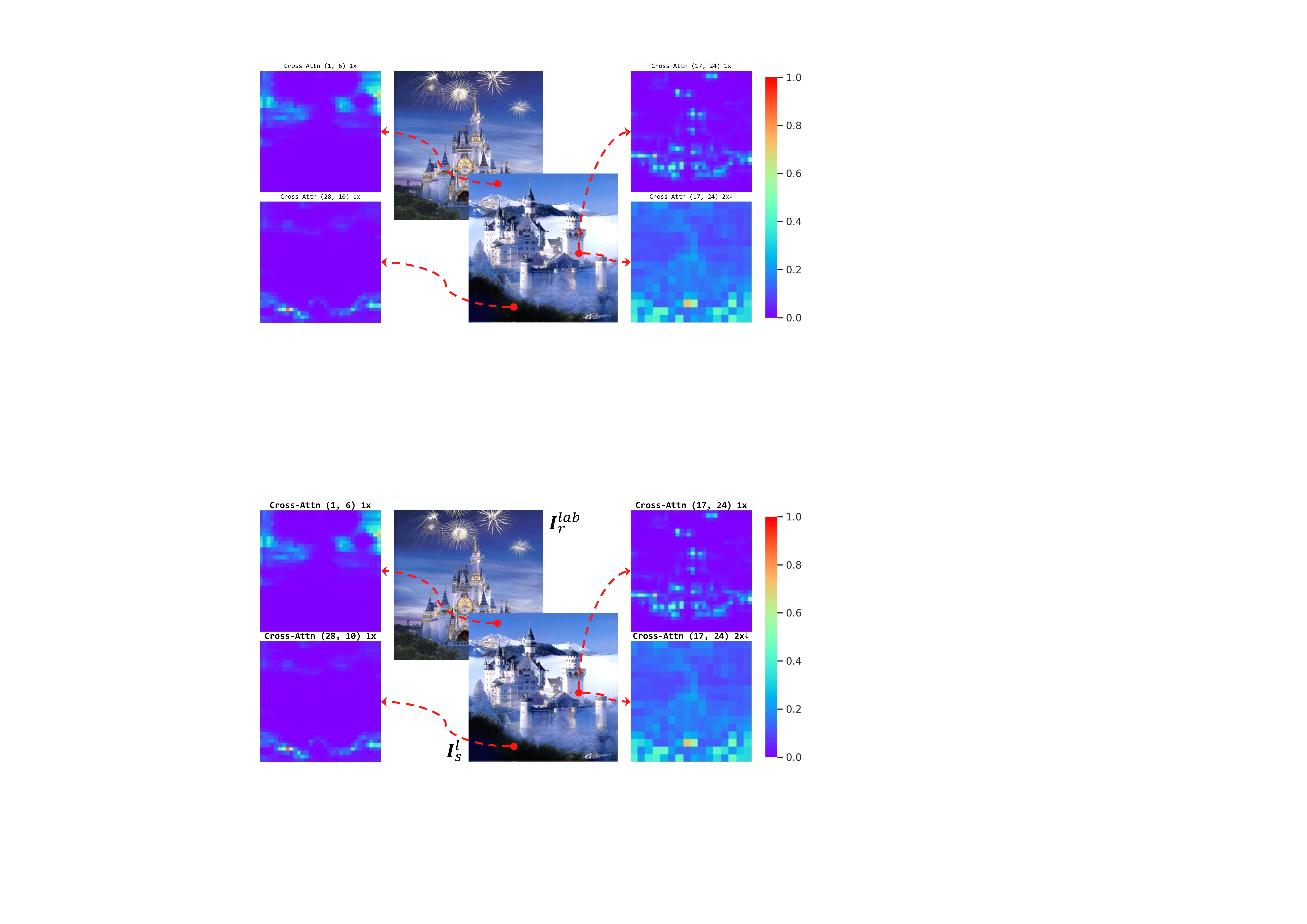}
   \caption{Cross attention for a set of reference points. The PVCAttn can match similar semantic information. The left part shows attention maps for different positions, while the right part shows pyramid attention maps for one position.}
   \label{fig:Vis}
\end{figure}

\begin{figure}[!tp]
   \centering
   \includegraphics[width=1.0\columnwidth]{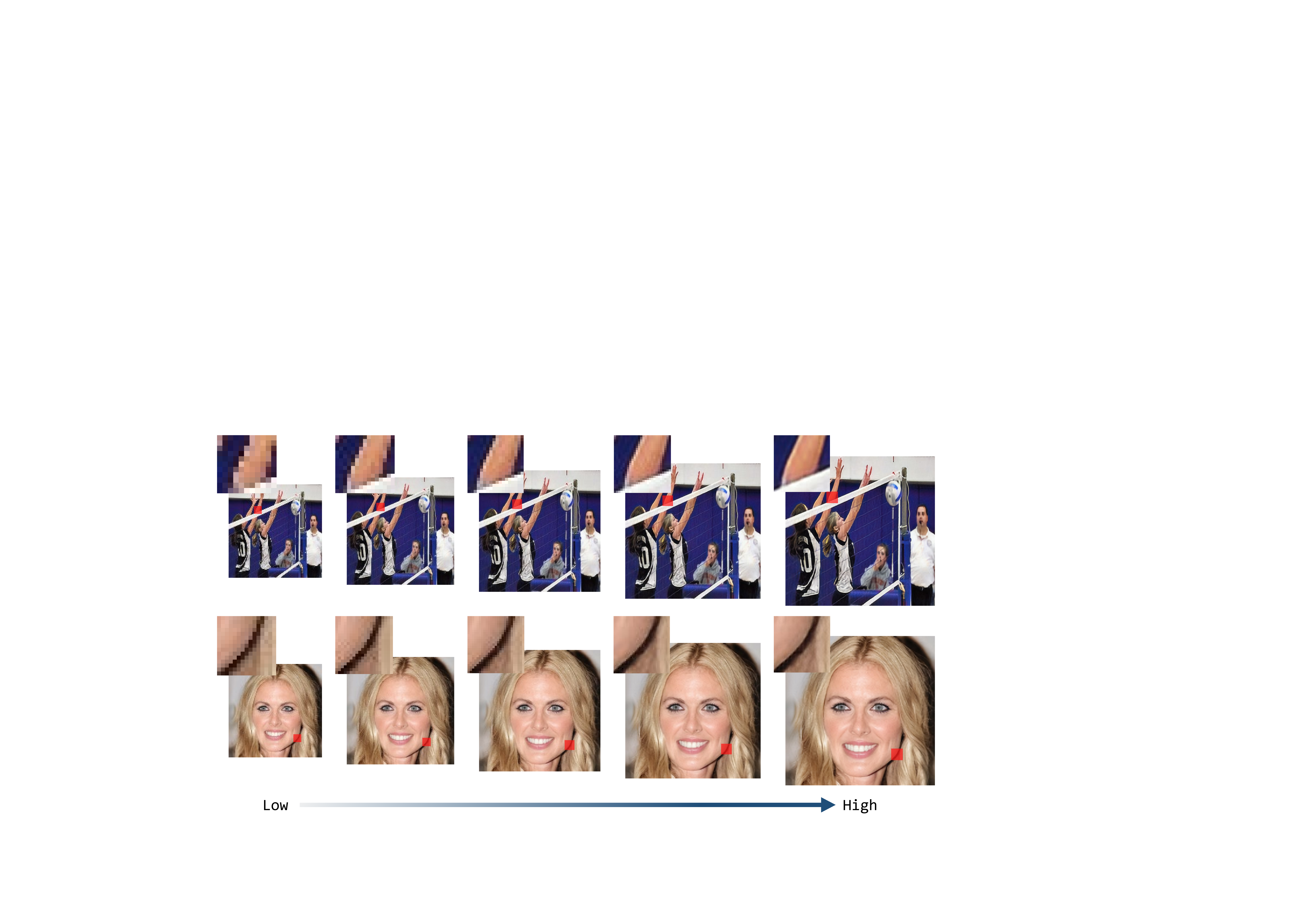}
   \caption{Image generation at continuous magnification. Magnified red areas are displayed in the upper left corner.}
   \label{fig:Interpolation}
\end{figure}

\noindent\textbf{Interpretability of PVCAttn.}
In Figure~\ref{fig:Vis}, we visualize the attention maps of PVCAttn, focusing on a few points in the source image. Visualized attention maps indicate that each location pays more attention to semantically similar areas, and the location in the low-resolution feature map focuses on more average areas (\cf right-bottom attention map).

\noindent\textbf{Multi-Magnification Generation.}
Benefit from the CPM module, SCSNet can generate target images at continuous magnification. As shown in Figure~\ref{fig:Interpolation}, the generated results have consistent color stability for different resolutions and smooth transitions for adjacent images. Dynamic video can be seen in the supplementary material.

\section{Conclusion}
In this paper, we propose an efficient paradigm to address SCS task and design an end-to-end SCSNet to complete this goal. Concretely, a \emph{PVCAttn} module is designed to aggregate feature information between source and reference images effectively, while the \emph{CPM} efficiently models the discrete pixel mapping in a continuous space to generate target images at arbitrary magnification. Extensive experiments demonstrate our approach's superiority for achieving the SCS task well and generating high-quality images. In the future, we will combine general detection and segmentation methods with current colorization branch to provide more semantic-knowability information. 

\section{Acknowledgments}
We thank all authors for their excellent contributions as well as anonymous reviewers and chairs for their constructive comments. This work is partially supported by the National Natural Science Foundation of China (NSFC) under Grant No. 61836015.

{
  \bibliography{main}
}

\end{document}